\documentclass[sigconf]{acmart}

\copyrightyear{2026}
\acmYear{2026}
\setcopyright{cc}
\setcctype{by}
\acmConference[WWW Companion '26] {Companion Proceedings of the ACM Web Conference 2026}{April 13--17, 2026}{Dubai, United Arab Emirates.}
\acmBooktitle{Companion Proceedings of the ACM Web Conference 2026 (WWW Companion '26), April 13--17, 2026, Dubai, United Arab Emirates}
\acmISBN{979-8-4007-2308-7/2026/04}
\acmDOI{10.1145/3774905.3793110}

\usepackage{hyperref}
\usepackage{xcolor}
\usepackage{pifont} 
\newcommand*\colourcheck[1]{%
\expandafter\newcommand\csname #1check\endcsname{\textcolor{#1}{\ding{52}}}%
}
\colourcheck{green}
\newcommand*\colourcross[1]{%
\expandafter\newcommand\csname #1cross\endcsname{\textcolor{#1}{\ding{56}}}%
}
\colourcross{red}

\usepackage{fontawesome5} 
\usepackage{graphicx}     
\usepackage{balance}
\usepackage[most]{tcolorbox}

\usepackage{enumitem}

\begin{document}

\title[Multimodal Peer Review Simulation with Actionable To-Do Recommendations]{Multimodal Peer Review Simulation with Actionable To-Do Recommendations for Community-Aware Manuscript Revisions}

\author{Mengze Hong}
\affiliation{%
  \institution{Hong Kong Polytechnic University}
  \city{Hong Kong}
  \country{China}
}

\author{Di Jiang}
\authornote{Corresponding Author}
\affiliation{%
  \institution{Hong Kong Polytechnic University}
  \city{Hong Kong}
  \country{China}
}

\author{Weiwei Zhao}
\affiliation{%
  \institution{AI Group, WeBank}
  \city{Shenzhen}
  \country{China}
}

\author{Yawen Li}
\affiliation{%
  \institution{Beijing University of Posts and Telecommunications}
  \city{Beijing}
  \country{China}
}

\author{Yihang Wang}
\author{Xinyuan Luo}
\affiliation{%
  \institution{Independent  Researcher}
  \city{Hong Kong}
  \country{China}
}

\author{Yanjie Sun}
\author{Chen Jason Zhang}
\affiliation{%
  \institution{Hong Kong Polytechnic University}
  \city{Hong Kong}
  \country{China}
}

\renewcommand{\shortauthors}{Mengze Hong et al.}

\newcommand\blfootnote[1]{%
  \begingroup
  \renewcommand\thefootnote{}\footnote{#1}%
  \addtocounter{footnote}{-1}%
  \endgroup
}

\begin{abstract}

While large language models (LLMs) offer promising capabilities for automating academic workflows, existing systems for academic peer review remain constrained by text-only inputs, limited contextual grounding, and a lack of actionable feedback. In this work, we present an interactive web-based system\blfootnote{\raisebox{-.01ex}{$\vcenter{\hbox{\includegraphics[height=1.3em]{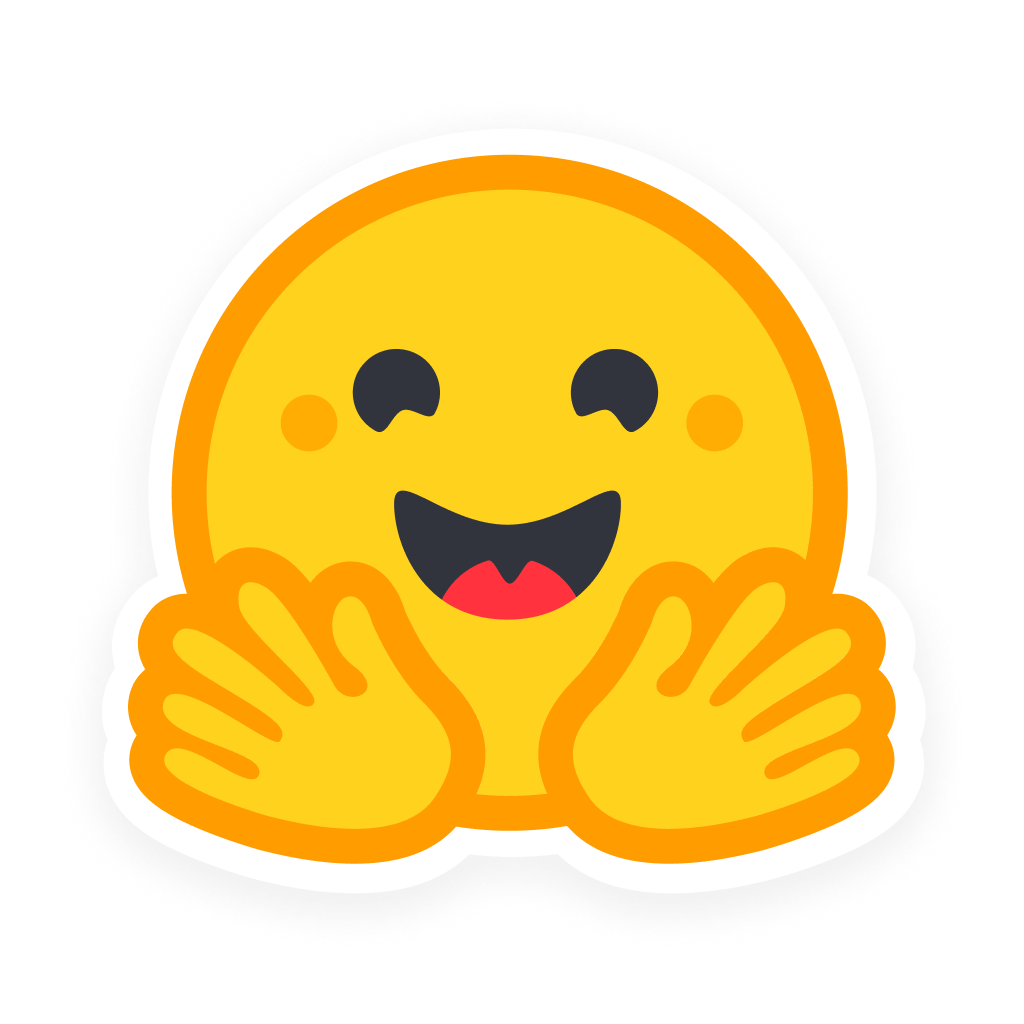}}}$}
Demo: \href{https://huggingface.co/spaces/mengze-hong/multimodal-peer-review-simulation}{huggingface.co/spaces/mengze-hong/multimodal-peer-review-simulation}\\ \hspace*{0.5pt} \faGithub\ \kern1pt Code: \href{https://github.com/mengze-hong/multimodal-peer-review-simulation}{github.com/mengze-hong/multimodal-peer-review-simulation}} for multimodal, community-aware peer review simulation to enable effective manuscript revisions before paper submission. Our framework integrates textual and visual information through multimodal LLMs, enhances review quality via retrieval-augmented generation (RAG) grounded in web-scale OpenReview data, and converts generated reviews into actionable to-do lists using the proposed Action:Objective[\#] format, providing structured and traceable guidance. The system integrates seamlessly into existing academic writing platforms, providing interactive interfaces for real-time feedback and revision tracking. Experimental results highlight the effectiveness of the proposed system in generating more comprehensive and useful reviews aligned with expert standards, surpassing ablated baselines and advancing transparent, human-centered scholarly assistance.

\end{abstract}



\maketitle

\section{Introduction}

\begin{table*}[ht]
    \centering
    \caption{Comparison of peer review generation systems.}
    \vspace{-0.5em}
    \resizebox{\textwidth}{!}{
    \begin{tabular}{lccccc}
        \toprule
        \textbf{Method} & \textbf{Text Summary} & \textbf{Multi-Dimensional} & \textbf{Actionable To-Do} & \textbf{Multimodal Perception} & \textbf{Web-Data Integration} \\
        \midrule
        MARG \cite{darcy2024margmultiagentreviewgeneration} & \greencheck & \greencheck & \redcross & \redcross & \redcross \\
        CycleResearcher \cite{wengcycleresearcher} & \greencheck & \greencheck & \redcross & \redcross & \redcross \\
        OpenReviewer \cite{idahl2025openreviewer} & \greencheck & \greencheck & \greencheck & \redcross & \redcross \\
        DeepReview \cite{zhu2025deepreviewimprovingllmbasedpaper} & \greencheck & \greencheck & \greencheck & \redcross & \redcross \\
        TreeReview \cite{chang2025treereview} & \greencheck & \greencheck & \redcross & \redcross & \redcross \\
        KID-review \cite{yuan2022kid} & \greencheck & \greencheck & \redcross & \redcross & \redcross \\
        SEAGraph \cite{yu2024seagraphunveilingstorypaper} & \greencheck & \greencheck & \redcross & \redcross & \greencheck \\
        \midrule
        \textbf{Ours} & \greencheck & \greencheck & \greencheck & \greencheck & \greencheck \\
        \bottomrule
    \end{tabular}}
    \label{tab:comparison}
\end{table*}

\begin{figure}[!t]
    \centering
    \includegraphics[width=\columnwidth]{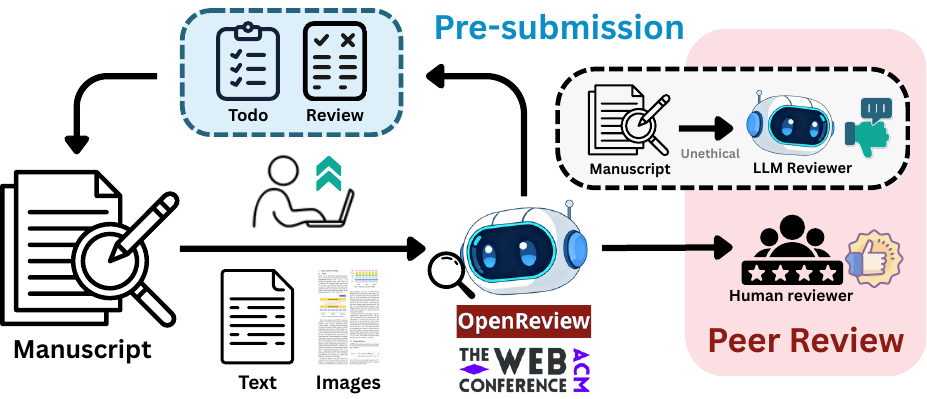}
    \vspace{-2em}
    \caption{Overview of the multimodal peer review simulation workflow for enhancing manuscripts before submission.}
    \Description[Overview of multimodal peer review simulation workflow.]{Diagram illustrating the workflow of a multimodal peer review simulation to enhance manuscripts before submission. The process begins with a research paper manuscript that typically contains text and images. Since it is unethical to submit the manuscript directly for peer review, this paper proposes performing a pre-submission peer-review simulation using an LLM. The workflow highlights interactions between automated LLM evaluation and human researchers to improve manuscript quality prior to submission to conferences or journals.}
    \label{fig:text}
    \vspace{-1em}
\end{figure}

The rapid advancement of large language models (LLMs) has reshaped many aspects of academic research and scholarly communication. These models have demonstrated the ability to process large volumes of information, summarize complex ideas, and provide consistent feedback at scale, positioning them as promising tools for scholarly tasks when supported by large-scale public research repositories (e.g., OpenReview) \cite{10.1145/3701716.3716888, 10.1145/3701716.3735083, 10.1145/3701716.3717752}. One emerging application is their use in the \textbf{peer review process} \cite{10.1145/3442442.3451370, 10.1145/3041021.3053053}, helping to manage the growing volume of paper submissions, accelerate decision-making, and support authors in improving their manuscripts. While substantial progress has been made toward LLM-based peer review simulation (see Table \ref{tab:comparison}), the current use of these systems within academic workflows raises serious concerns\footnote{See \textbf{Non-LLM Policy for Reviewing} in
\href{https://www2026.thewebconf.org/calls/industry.html}{TheWebConf}, \href{https://icml.cc/Conferences/2026/LLM-Policy}{ICML}, \href{https://neurips.cc/Conferences/2025/LLM}{NeurIPS}, and \href{https://blog.iclr.cc/2025/11/19/iclr-2026-response-to-llm-generated-papers-and-reviews/}{ICLR}.}
and critical limitations that caution against inappropriate scholarly evaluation:

\begin{itemize}
    \item \textbf{Unethical}: Reliance on LLMs to produce entire reviews during the formal peer review process bypasses the intellectual engagement and critical thinking essential to reviewers, undermining the integrity of scholarly evaluation.
    \item \textbf{Unreliable}: Existing LLM-based review generation relies primarily on textual input, ignoring crucial visual elements such as figures, tables, and layout. Furthermore, there is a lack of understanding of the varying peer review standards, resulting in misleading feedback.
    \item \textbf{Unactionable}: Current approaches emphasize providing comments and criticisms without offering concrete and targeted guidance for improvement, making it difficult for authors to incorporate the feedback into meaningful revisions.
\end{itemize}

Recently, the concept of pre-submission review assistance for authors has emerged \cite{idahl2025openreviewer}, promoting the ethical integration of LLMs into scholarly workflows, similar to paper proofreading. However, relying on text-only inputs introduces bias, as human reviewers form initial impressions based on the visual structure, figures, and tables before considering textual details. Prior works have explicitly acknowledged the lack of consideration for multimodal input as a core limitation \cite{chang2025treereview, wengcycleresearcher}, underscoring the need for system innovations. To bridge these gaps, we present a novel system designed to help human authors improve their manuscripts through comprehensive feedback generated by multimodal peer review simulation. The system's key innovations include: (1) the integration of multimodal LLMs that jointly process textual and visual content, (2) the generation of actionable, structured to-do suggestions in the \texttt{Action:Objective[\#]} format, and (3) community-aware review generation informed by retrieval-augmented generation (RAG) from web-scale peer review data in the OpenReview repository, ensuring that feedback aligns with established academic standards.

The main contributions of this work are as follows:

\begin{itemize}
\item We present the first open-source web demo system that integrates multimodal LLMs to deliver human-aligned review feedback, grounded in publicly available review datasets from OpenReview for the target venue.
\item We propose a structured and actionable to-do format that distills reviewer commentary into traceable revision items, enabling authors to systematically apply feedback for manuscript improvements before submission.
\item Empirical results demonstrate the critical role of multimodal perception in producing more comprehensive and balanced assessments, reducing bias, and enhancing feedback reliability and usefulness in academic workflows.
\end{itemize}

\begin{figure*}
    \centering
    \includegraphics[width=0.75\linewidth]{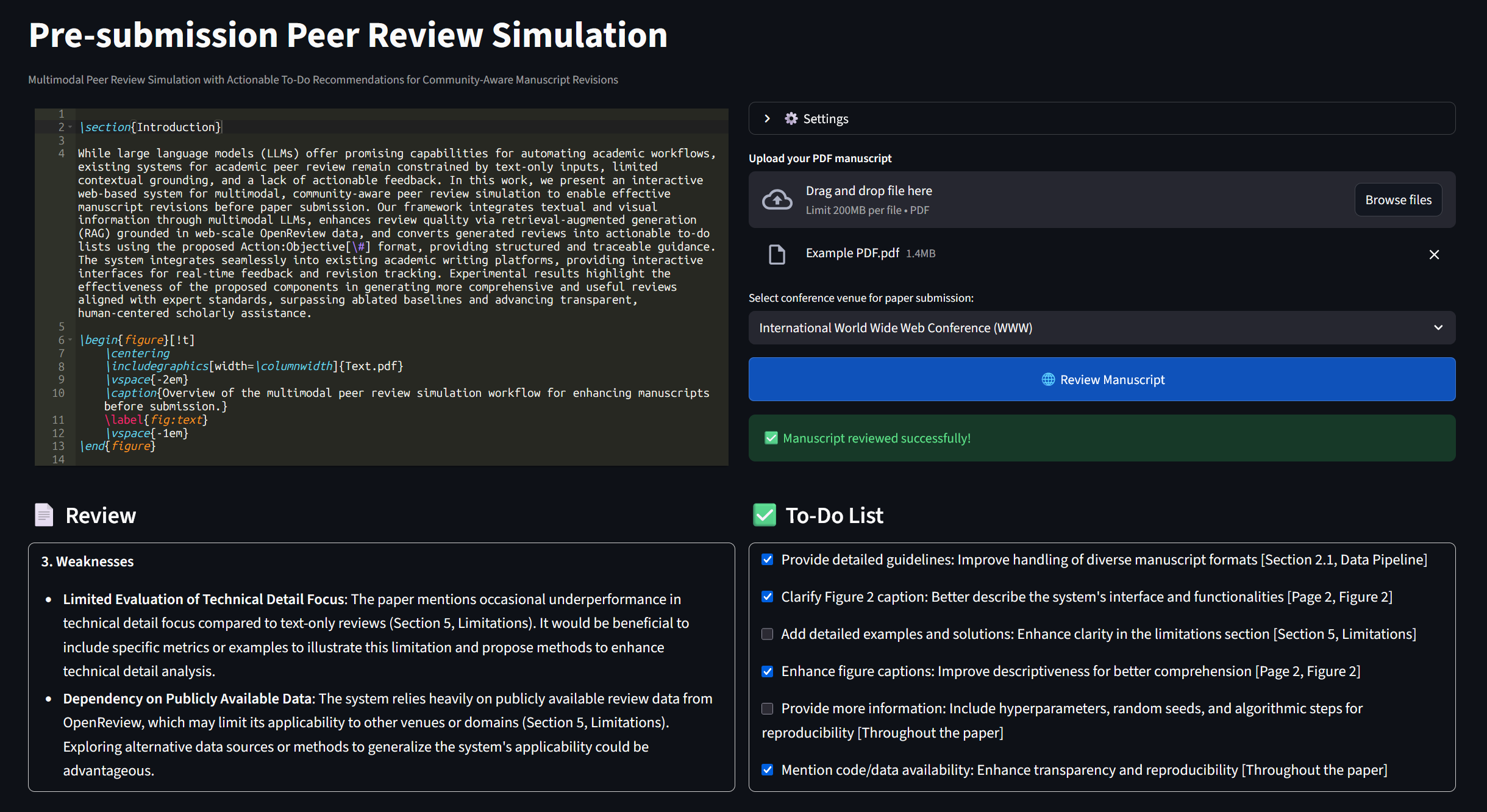}
    \caption{Illustration of the peer review simulation system. The top-left panel displays the LaTeX editor, while the right panel outlines the system workflow, including LLM API configuration, PDF upload, and selection of the submission venue. Review outcomes are displayed in Markdown at the bottom, with to-do checkboxes that let users tick the ready-made revisions.}
    \Description[A screenshot of the proposed multimodal peer review simulation system.]{Screenshot of a web-based multimodal peer review simulation interface. The left side shows a LaTeX source code editor with paper text. The right side contains: a PDF upload area, a conference venue dropdown (WWW selected), a "Review Manuscript" button, a success message, and a structured review panel displaying weaknesses with actionable to-do checkboxes. The bottom section presents review feedback in markdown format with checklist items for manuscript improvement.}
    \label{fig:system_illustration}
\end{figure*}

\section{Proposed System}

We propose an end-to-end framework for multimodal peer review simulation, designed to help authors improve their manuscripts prior to submission to a target venue. The system accepts a user-uploaded PDF and operates through two primary stages: data preprocessing and multimodal LLM integration.

\subsection{Data Pipeline}

In the data processing stage, textual and visual components of the input PDF are processed in parallel to facilitate multimodal inference. For the textual stream, the raw text content is extracted via PDF parsing, with OCR applied when necessary. To address input-token constraints that limit the processing of long manuscripts, we employ hierarchical summarization \cite{christensen-etal-2014-hierarchical} to construct a structured, multi-level representation that preserves essential information while compressing content. In particular, sentences are aligned to sections via dependency-parsing heuristics, adapting temporal clustering to match the structural conventions of research papers. We apply a recursive top-down clustering approach that partitions the text into clusters \(C = \{c_1, \dots, c_k\}\) by optimizing  
\[
\max_C B(C) + \alpha E(C),
\]
where \(B(C)\) detects content-density bursts for determining breakpoints, and \(E(C)\) ensures balanced cluster sizes; \(\alpha\) is tuned empirically. This hierarchical representation captures the manuscript’s logical flow from general overview to detailed discussion, ensuring minimal information loss.  

For the visual stream, each page of the PDF is rendered as a high-resolution PNG image to preserve graphical elements such as figures, tables, and layouts. Since research papers often exceed 20 pages, we concatenate every four consecutive pages into a single composite image. This reduces computational overhead while preserving spatial continuity across the document, enabling efficient global visual reasoning and consistent positional alignment across pages. To balance visual fidelity and computational cost, all images are normalized to a fixed resolution, and page boundaries are retained as separators to aid layout interpretation. This design facilitates cross-page context recognition, enabling the MLLM to better correlate visual elements with corresponding textual regions and thereby improve the overall coherence of multimodal inference.

\subsection{Multimodal LLM Processing}

We integrate multimodal large language models with web-scale review datasets drawn from OpenReview to generate structured, traceable reviews and actionable to-do feedback.

\subsubsection{Prompt Design}
The condensed hierarchical text and concatenated images are jointly input to a multimodal LLM (e.g., GPT-4o or DeepSeek for local deployment). The base prompt is designed in accordance with established principles of effective peer review \cite{jefferson2002measuring} and the ICLR 2025 reviewer guidelines, emphasizing originality, soundness, clarity, and significance. To leverage visual input, the model is explicitly instructed to engage with figures and tables. To better align with human reviewers, we perform prompt tuning to iteratively refine the base prompt using over 1000 review samples from OpenReview. The LLM first generates candidate reviews \(\hat{r}\) from the initial prompt, which are evaluated against human review references \(r\) using a composite metric: 
\[
S = \lambda_1 \cdot \mathrm{ROUGE}(\hat{r}, r) + \lambda_2 \cdot \mathrm{BERTScore}(\hat{r}, r),
\]  
and used in a ``\textit{critique–refine}'' process to adjust the prompt’s wording and emphasis, so that subsequent generations better mirror the depth, preferences, and structure of human reviewers.

\subsubsection{RAG with OpenReview Datasets} 

To enforce consistent alignment with community standards, we employ RAG to incorporate insights from publicly accessible review corpora. For a user-uploaded manuscript, the title and abstract are jointly embedded, after which the system retrieves the top-2 most relevant research papers from the OpenReview repository of the target venue using cosine similarity. The corresponding human review data of these papers are summarized into key aspects of review practices, abstracting away paper-specific content while emphasizing structural conventions, evaluation dimensions, and common judgment patterns of the target venue, as reflected in papers of similar type and topic. This information is inserted into the review generation process as contextual guidance, which naturally steers the LLM to produce structured outputs that mirror real-world scholarly feedback specific to the target venue, thereby fostering community-aware and contextually grounded reviews. Finally, each generated review item is required to include explicit references to page or section numbers (e.g., ``Page 5'', ``Figure 2''), ensuring precise localization within the original PDF.

\subsubsection{To-Do Generation}

The goal of peer review simulation at the pre-submission stage is to provide actionable insights for manuscript revision, which can be effectively achieved through a list of to-do suggestions that authors can easily track. We propose a novel structured format for converting generated reviews into actionable to-do items, denoted as \texttt{\textbf{Action:Objective[\#]}}, where \textbf{Action} represents a specific, executable instruction (e.g., ``Revise figure caption''), \textbf{Objective} offers a concise rationale explaining the purpose of the action to aid author understanding, and \textbf{[\#]} refers to a precise locator within the manuscript, such as a section number. This structure ensures each feedback item is concrete, explainable, and traceable throughout the document. For example:

\begin{tcolorbox}[
    colback=gray!10,
    colframe=gray!60,
    boxrule=0.5pt,
    arc=2pt,
    left=7pt,
    right=7pt,
    top=2pt,
    bottom=2pt
]
\begin{itemize}[leftmargin=1.2em, labelsep=0.8em, itemsep=0.2em]
    \item[$\boxtimes$] \texttt{Revise introduction: Describe the research gaps [Section 1. L12--L18]}
    \item[$\square$] \texttt{Update figure caption: Improve interpretability with detailed descriptions [Page 5. Figure 3]}
    \item[$\square$] \texttt{Add citation: Ensure academic rigor for metric selections [Section 4.1]}
\end{itemize}
\end{tcolorbox}

\subsection{Demo Interface}
The proposed system prioritizes user experience by embedding the peer review simulation as a core utility within the broader scholarly workflow of academic writing. The system provides manuscript editing functionalities comparable to standard LaTeX editors, with the review simulation available as an on-demand feature directly in the interface. Built using the Streamlit library
, as illustrated in Figure \ref{fig:system_illustration}, the system allows users to upload a PDF file or to compose a paper in the integrated LaTeX editor. The review process is optimized to ensure minimal latency, and the generated review is rendered in markdown format, followed by a comprehensive, actionable to-do list presented in checkboxes, enabling authors to track revisions systematically. For practical deployment, the system utilizes locally deployed models such as DeepSeek-R1 or Llama-3.2-90B-Vision-Instruct to ensure the privacy of unpublished research articles. For demonstration purposes, however, the third-party OpenAI API is employed to simplify model configuration and minimize computational resource requirements.

\begin{figure}
    \centering
    \includegraphics[width=1\linewidth]{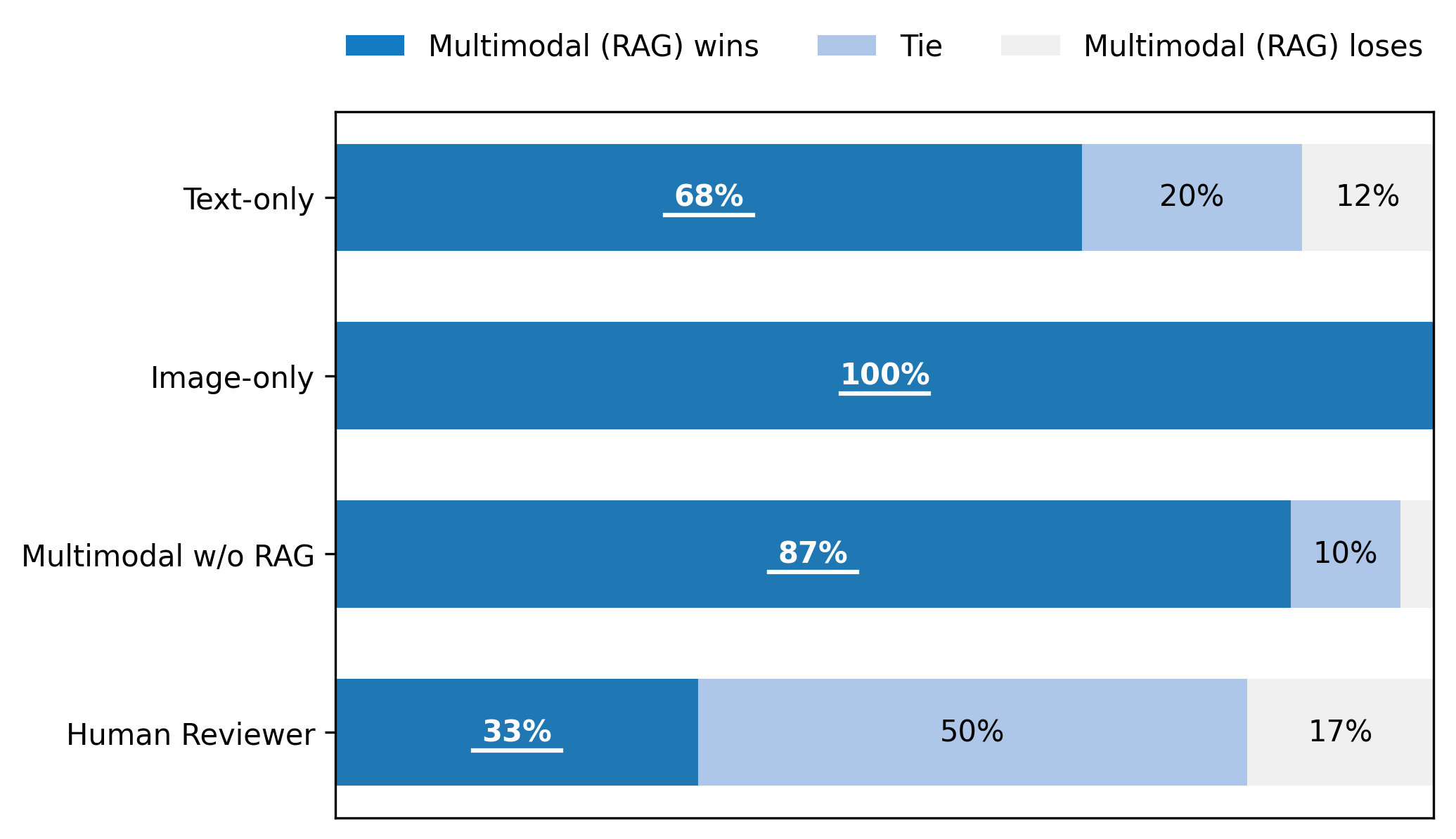}
    \vspace{-2em}
    \caption{Human preference evaluation of review quality across generation approaches. Percentages \underline{underlined} indicate the proportion of cases where the proposed Multimodal (RAG) system outperforms the baseline.}
    \Description[Figure of experimental results.]{Horizontal bar chart showing human preference evaluation of review quality. Categories (y-axis): Text-only, Image-only, Multimodal w/o RAG, Human Reviewer. Bars segmented into blue (Multimodal with RAG wins), light blue (Tie), gray (Multimodal with RAG loses).}
    \vspace{-1em}
    \label{fig:performance}
\end{figure}

\section{Experiments}

To evaluate the effectiveness of proposed methods, we compare our system against three ablated baselines: (1) text-only input, (2) image-only input, and (3) multimodal input without RAG integration. All methods utilize GPT-4o for its strong visual understanding capabilities. Evaluation is performed using both human assessments with 5 PhD-level researchers and LLM-as-a-judge protocols based on GPT-5, determining whether our system outperforms, matches, or underperforms each baseline in terms of review quality. The evaluation dataset comprises 30 randomly selected ICLR 2023 papers, each accompanied by the corresponding human reviews as the reference ground-truth. To support retrieval, we construct a RAG database from 1,000 ICLR submissions in 2021--2022, where each paper’s title and abstract are combined into a single embedding using SentenceTransformer to enable efficient data retrieval.

\paragraph{Review Quality} Evaluation results show that the multimodal RAG approach significantly outperforms all ablated baselines under human evaluations (see Figure \ref{fig:performance}. Specifically, it secures wins in 68\% of text-only comparisons, 100\% against image-only inputs, and 87\% versus multimodal setups without RAG. In the few cases where it underperforms the text-only approach, we find that purely textual reviews tend to focus more on technical details, whereas the multimodal approach provides a more balanced and comprehensive assessment. This suggests that the selection of suitable review strategies may also depend on the nature and presentation of the manuscript. Human evaluations place our method on par with ground-truth human reviewers, with 33\% wins and 50\% ties. Moreover, the GPT-5 judgments favor the proposed method in winning \textbf{72\%} of cases against human reviews, likely reflecting documented biases toward machine-generated content and underscoring the need for human involvement in academic evaluation and judgment.

\noindent \paragraph{Usefulness} We further evaluate the usefulness of our peer review simulation by comparing the proposed to-do list feedback with traditional plain text comments. Five human researchers assessed reviews for 30 research papers, annotating the number of actionable improvements derived from each review type. Results show that the to‑do list feedback increases the number of actionable items by \textbf{57.6\%}. On average, the to‑do approach yields five actionable items, compared to only two in plain reviews, which can be attributed to the clear action structure and reference locator in the proposed to-do format, carrying practical benefits in academic writing. 

\section{Conclusion}

This paper presented a multimodal, community-aware peer review simulation system that integrates textual and visual understanding through multimodal LLMs, enhances feedback quality via RAG grounded in web-scale OpenReview data, and provides interpretable, actionable revision recommendations to support pre-submission manuscript improvement. Experimental results demonstrate the system’s effectiveness in producing comprehensive, expert-aligned reviews, highlighting its potential as a transparent, human-centered tool for scholarly workflows. Despite its strong performance, several limitations remain, including the reliance on a limited set of publicly available review data and the dependence on OCR-based document parsing, which could be improved by incorporating raw LaTeX sources directly. Future work is encouraged to build upon the released project, enhancing contextual grounding through dynamic retrieval and incorporating interactive dialogue communication to maintain alignment with evolving user demands.

\begin{acks}
The work was partially supported by the PolyU Start-up Fund (P0059983), the NSFC/RGC Joint Research Scheme (N\_PolyU5179/25), the National Natural Science Foundation of China (62532002), and the Research Grants Council (Hong Kong) (PolyU25600624).
\end{acks}

\bibliographystyle{ACM-Reference-Format}
\balance
\bibliography{references}

\end{document}